# Pig behavior dataset and Spatial-temporal perception and enhancement networks based on the attention mechanism for pig behavior recognition

Fangzheng Qi[a], Zhenjie Hou[a,*], En Lin[b], Xing Li[c], Jiuzhen Liang[a], Xinwen Zhou[a]

[a]School of Computer Science and Artificial Intelligence, Changzhou University, Changzhou, 21300, China

[b]Goldcard Smart Group Co., Ltd, Hangzhou, Zhejiang 310018.China

[c]College of Information Science and Technology, College of Artificial intelligence, Nanjing Forestry University, Nanjing, 210098, China

## Abstract

The recognition of pig behavior plays a crucial role in smart farming and welfare assurance for pigs. Currently, in the field of pig behavior recognition, the lack of publicly available behavioral datasets not only limits the development of innovative algorithms but also hampers model robustness and algorithm optimization.This paper proposes a dataset containing 13 pig behaviors that significantly impact welfare.Based on this dataset, this paper proposes a spatial-temporal perception and enhancement networks based on the attention mechanism to model the spatiotemporal features of pig behaviors and their associated interaction areas in video data. The network is composed of a spatiotemporal perception network and a spatiotemporal feature enhancement network. The spatiotemporal perception network is responsible for establishing connections between the pigs and the key regions of their behaviors in the video data. The spatiotemporal feature enhancement network further strengthens the important spatial features of individual pigs and captures the long-term dependencies of the spatiotemporal features of individual behaviors by remodeling these connections, thereby enhancing the model's perception of spatiotemporal changes in pig behaviors. Experimental results demonstrate that on the dataset established in this paper, our proposed model achieves a MAP score of 75.92%, which is an 8.17% improvement over the best-performing traditional model. This study not only improces the accuracy and generalizability of individual pig behavior recognition but also provides new technological tools for modern smart farming. The dataset and related code will be made publicly available alongside this paper.

---

* Corresponding author E-mail address: houzj@cczu.edu.cn (Z. Hou).

# 1. Introduction

  Pigs are among the most commonly raised livestock worldwide. Their rich content of protein, vitamins, and fats, which are beneficial to human health, has made them one of the most widely consumed livestock globally. With the continuous development in intensive farming, the health and welfare of pigs have become a focal point of concern, as they are crucial for reducing animal stress and improving the quality of livestock products(Ma et al., 2023[1]; Mellor, 2016[2]; Pedersen, 2018[3]; Liu et al., 2020[4]). For example, defining different lying patterns based on Delaunay triangulation (DT) features extracted from pig group lying patterns can help farm managers assess whether the thermal comfort provided to pigs in large farms is adequate (Nasirahmadi et al., 2017[5]). Additionally, pigs' reactions to new objects are intrinsically linked to their early-life dietary patterns, indicating that environmental factors can influence brain development. Consequently, assessing new object recognition in pigs can provide insights into their memory and feeding costs (Shirke et al., 2021[6]). Detecting and recognizing the behavior of pigs and other animals requires addressing the following two main issues: (1) the development of relevant datasets. High-quality datasets are crucial for training effective deep learning models, with the scale, quality, and diversity of the dataset directly affecting model accuracy and generalization; (2) the design of detection and recognition algorithms.

  The construction of animal behavior datasets, particularly for farm animals, has increasingly attracted the attention of researchers.For example, Kashiha et al. (2013) [7] combined video surveillance with dynamic modeling to monitor the drinking behavior of pig groups using water meters and CCD cameras, achieving an accuracy rate of 92%. This provided effective support for estimating water consumption. However, the dataset primarily focused on drinking behavior, lacking diversified support for other behaviors, and its generalization ability in complex scenarios has not been verified. Kim et al. (2021) [8] proposed a real-time feeding behavior detection method based on deep learning, optimizing performance with the YOLO model, achieving a feeding behavior recognition accuracy of 95.66% to 96.56%. This method improved detection speed by merging pig detection and behavior classification tasks, but it was also limited to feeding behavior, did not consider other behavior categories, and lacked cross-scenario adaptability.Yang et al. (2018) [9] employed Faster R-CNN to locate and identify individual pigs in group-housed environments and achieved behavior recognition based on feeding area occupancy. However, their annotations were primarily image-based, focusing on the localization of pig heads and bodies, with behavior recognition relying on spatial features. This approach lacked direct video-based behavior category annotations, limiting the recognition of dynamic behaviors such as playing and social interactions. Additionally, Gao et al. (2023) [10] released a pig aggression behavior dataset that annotated aggressive actions like biting, pushing, trampling, and chasing, focusing on interaction behaviors. However, this dataset does not meet the requirements of practical production environments. Shirke et al. (2021) [11] introduced a multi-camera pig activity dataset with global IDs for each pig, addressing cross-camera perspective transitions. Nevertheless, its behavior annotations were relatively coarse and did not encompass complex behavioral patterns. Furthermore, Shirke et al. (2021) [11] also provided a novel preference behavior dataset supporting action recognition and keypoint detection. Despite its flexible annotation format, the dataset still exhibited limitations in terms of behavioral category and scene diversity.To address these issues, we developed a video-based pig behavior dataset with significant implications for pig welfare. This dataset provides detailed annotations of various pig behaviors during rearing, with a particular focus on welfare-related behaviors. For instance, it includes annotations for behaviors such as standing up and lying down, which are critical for studying whether sows cause harm to piglets during lactation, and movement and running behaviors, which help assess whether group-housed pigs achieve the expected level of physical activity. These annotations offer valuable data support for further analysis of pig health and behavioral patterns.

Traditional livestock monitoring methods are time-consuming, costly, disruptive to farm management processes, and reliant on subjective judgment, posing significant challenges in large-scale pig production. With advances in automation equipment, image processing, and deep learning, animal behavior monitoring has transitioned from manual observation to automated detection. Video-based deep learning is increasingly being utilized on a large scale to detect and recognize various pig behaviors. This approach not only enhances the accuracy of animal behavior recognition but also minimizes the influence of subjective human factors on detection results, thereby improving both precision and robustness. Yang et al. (2020) [12] reviewed the recent advancements in automated video-based pig behavior recognition algorithms, highlighting that technologies such as pig segmentation, pig detection, and behavior recognition have become key research focuses in the field of intelligent animal husbandry.

Pig segmentation aims to obtain the complete contour of the monitored subject in a given image or specific video frame or to generate a series of key points representing their outline. This serves as the foundation for analyzing various welfare indicators of pigs, such as body size and weight. In early pig segmentation practices, segmentation was achieved by extracting binary contours of pigs from the background. For instance, Nasirahmadi et al. (2019) [13] and Zhu et al. (2015) [14] implemented pixel-level pig segmentation in static images using threshold segmentation algorithms and watershed transformations.In video data, traditional threshold segmentation methods often lead to confusion among multiple segmentation targets. Therefore, inter-frame differences are introduced in dynamic image segmentation to detect foreground objects. Guo et al. (2014) [15] utilized a Gaussian Mixture Model (GMM) for foreground extraction and combined it with maximum entropy threshold segmentation to ultimately obtain pig contours.Individual pig detection is critical for precision livestock farming. It involves identifying the potential locations of specific targets within an image and recognizing them accurately.Video-based pig detection and tracking face three major challenges.(1)Lighting variations: Changes in lighting conditions, such as differences between daytime and nighttime, can alter the camera imaging model.(2)Similar appearances: Pigs have highly similar appearances, and the background may exhibit varying states, further complicating detection.(3)Body deformation and occlusion: Pigs often gather closely, leading to body deformation and mutual occlusion.In the last case, although individual pig detection and tracking are considered critical stages for many video-based pig monitoring applications, developing a robust multi-target pig detection system remains a significant challenge (Zhang et al., 2019 [16]).Video-based pig behavior recognition is primarily achieved using image processing techniques, machine learning, and deep learning methods. Viazzi et al. (2014) [17] extracted two features—average motion intensity and occupancy index—from segmented regions of motion history images. By linking these features to aggressive interactions, they classified aggressive events using linear discriminant analysis. Thermal imaging (IRT) data obtained via infrared sensors have also been used to measure temperature distribution. For example, Amezcua et al. (2014) [18] identified lameness in sows by analyzing their average leg temperature, while Scolari et al. (2011) [19] compared changes in vulvar and rump skin temperatures to detect estrus behavior in sows.Unlike methods that rely solely on spatial features from video frames to recognize pig behaviors, Gao et al. (2023) [10] proposed a hybrid model combining convolutional neural networks (CNN) and gated recurrent units (GRU) to differentiate aggressive behaviors from other behaviors in videos. In this model, CNNs extract spatial features representing the motion-related featuresof each video frame, while GRUs capture temporal features to model motion patterns over time.Building on spatiotemporal features extracted from pig video data, Chen et al. (2020) [20] used the InceptionV3 model to extract spatial features from individual video frames. These features were then fed into a long short-term memory (LSTM) framework to model the spatiotemporal relationships between pigs and objects of interest. This approach was used to identify enriched engagement (EE) behaviors and preliminarily determine object preferences, enabling the prevention of tail-biting and aggressive behaviors during rearing.

Although deep learning-based feature learning methods have achieved remarkable progress in animal behavior recognition, several challenges still limit their application in practical pig farming scenarios. For instance, Yang et al. (2021) [21] developed a pig detector based on Faster R-CNN to locate the body, head, and tail of pigs in image frames. They then selected spatial features such as the distance between two piglets, overlapping area, and intersection angle within a single frame to classify mounting and non-mounting behaviors using an XGBoost classifier. Similarly, Li et al. (2023) [22] introduced a dual-stream deep neural network model to extract spatiotemporal features for the preliminary recognition of sow nursing behaviors. Liu et al. (2020) [23] employed detection and tracking algorithms to simplify group-level behaviors into interactions between two individuals. By combining convolutional neural networks (CNNs) with recurrent neural networks (RNNs), they extracted spatiotemporal features to classify tail-biting behaviors.Most traditional pig behavior recognition algorithms focus on specific behaviors, resulting in models with limited robustness in practical farming scenarios. Additionally, these studies often rely on manually designed features for machine learning or focus on a single behavior. When expanding farming operations or introducing new behavior classification requirements, these models often require re-annotating data and retraining, making them challenging to deploy immediately in production.To enhance the generalizability of behavior classification models across various scenarios, this study focuses on pig behavior recognition. Based on the constructed dataset, we propose a novel pig behavior classification algorithm designed to address these limitations and improve practical applicability.

In pig behavior classification tasks, traditional classification models often fail to explicitly model the key interaction regions between pigs and their surroundings. The extracted features for classification typically encompass all spatial data, thereby diminishing the importance of critical regions and increasing unnecessary computational costs. This limitation becomes more pronounced in large-scale farms, where the features decisive for behavior prediction are often associated with a small subset of spatial and temporal data relevant to the target behavior.For instance, in a video containing multiple pigs engaged in different activities within the same space, such as one pig drinking water while another plays, close proximity between the two can lead to reduced recognition accuracy and increased model training costs. Inspired by the Actor-Centric Relation Network (ACRN) model for behavior detection, constructing a relation network based on subject-centric feature representation to capture the interactions between the subject and other elements in the background can effectively model the spatiotemporal dynamics of pigs and their activity regions.

In this study, based on the constructed dataset, we propose an spatial-temporal perception and enhancement networks based on the attention mechanism to try to model the different interaction regions involved in the different behaviors of pigs and pigs, finally realizing the identification classification. This network consists of two branches with different sampling rates, comprising a spatiotemporal perception network and a spatiotemporal feature enhancement network. The spatiotemporal perception network is responsible for capturing the critical spatial features of individual pig behaviors and the key motion features at specific positions during movement, removing extraneous information unrelated to the target pig. The spatiotemporal feature enhancement network further strengthens the extracted spatiotemporal feature map of each pig by utilizing a channel-level spatial attention mechanism and a motion feature enhancement module. Through these components, the model's focus on behavioral variations is improved, ultimately achieving accurate recognition and classification of welfare-related pig behaviors.

Overall, the contributions is as follows:

This paper presents a video dataset comprising 13 pig behaviors that significantly impact pig welfare. The dataset provides detailed annotations of various behaviors observed during rearing, allowing for the study of welfare-related actions such as standing and lying behaviors to assess potential harm to piglets during sow nursing, and movement and running behaviors to evaluate if group pigs achieve adequate exercise. Researchers can use this dataset to create tailored subsets for exploring specific aspects of pig welfare.

This study proposes a Spatiotemporal Perception Network (STPN) designed to extract critical spatial features of target pigs and capture subtle variations in key behavioral features from video data.In the low-frame-rate branch, the spatiotemporal perception network uses a    frame-level spatial attention residual network to extract critical spatial features of individual pig behaviors. In the high-frame-rate branch, it builds on these critical spatial features and employs an inter-frame difference residual network to capture motion features at key positions during pig movement, effectively removing redundant information unrelated to individual pigs.The Spatiotemporal Feature Enhancement Network further enhances the individual pig's important spatial features using a channel-level spatial attention mechanism and a feature enhancement module.It aggregates spatial and temporal contexts to capture the spatiotemporal dependencies of individual pig behavior changes over time, enhancing the model's focus on different behavioral variations in pigs. This ultimately allows the successful modeling of the relationship between pigs and their key behavioral regions, achieving accurate classification of pig behaviors.

## 2. Materials and methods

## 2.1 Dataset intorduction

Based on the pig video dataset publicly released by Bergamini et al. (2021) [24], additional annotations were made to the unlabeled data for the purposes of this study.The original dataset contains 1,783 one-minute MP4 video clips.This dataset includes 8 pigs with special markings on their backs and primarily studies 5 types of behaviors: standing, lying down, moving, eating, and drinking. Ultimately, only 10 video clips were annotated.The data collection environment, depicted in Fig.1, has a 3-compartment feeding trough for free feeding, two nipple drinkers, and a plastic enrichment device mounted on the wall. The floor is covered with straw and shredded paper, with some areas consisting of slatted flooring. Data was collected using an Intel RealSense D435i camera positioned 2.5 meters above the ground, capturing RGB images at a frame rate of 6 fps and a resolution of 1280 x 720 pixels. The images were stored in batches of 1,800 frames (5 minutes) and compiled into one-minute videos with a frame rate of 30 fps. This dataset provides frame-level annotations for the five common pig behaviors. the original dataset provides frame-level annotations for the collected video data, marking only five common pig behaviors. The amount of annotated data is limited, and the range of behavior types is not extensive.

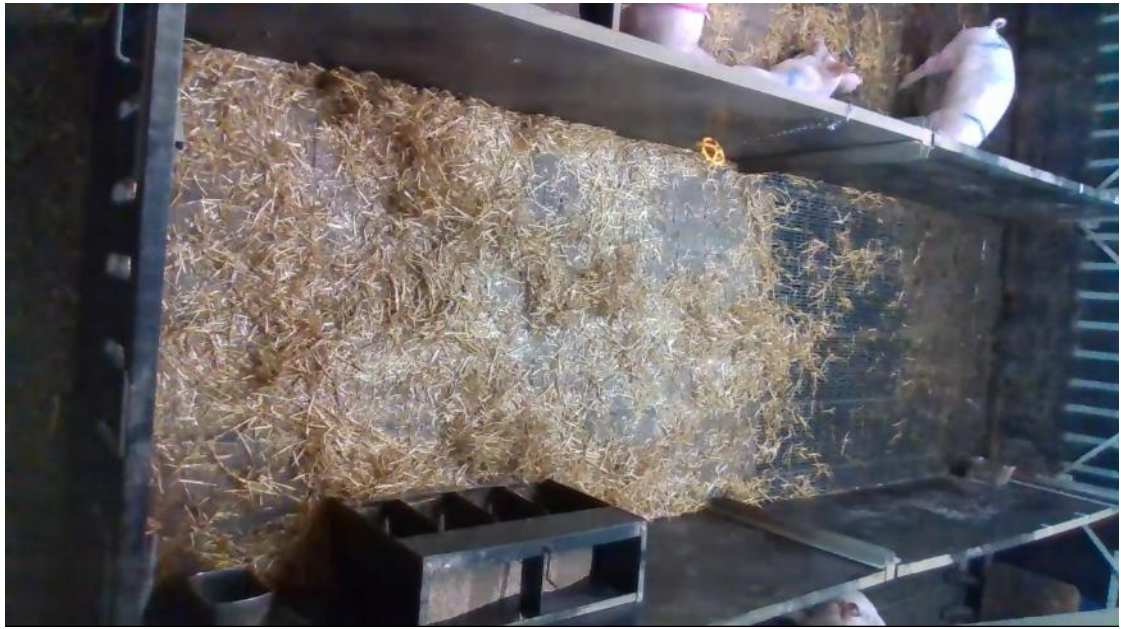

**Figure** 1.Pig activity data collection environment.

After analyzing the original dataset and publicly available pig behavior recognition datasets, we found that most accessible datasets focus on specific pig behaviors, resulting in trained models with insufficient robustness for actual feeding processes. These models are unable to guide other welfare-related production activities effectively, making it challenging for farmers to make balanced decisions between pig welfare and production costs. To facilitate subsequent research, we have summarized various pig behavior studies from recent years. Based on the construction method of the AVA dataset, we have reorganized and annotated 13 behaviors that significantly impact pig welfare. This is intended to enhance the generalizability of the trained model in practical production settings. In addition, to model the relationship between pigs in different behavioral states and their interaction regions, we have placed greater emphasis on the distinctions between behaviors, especially for overlapping behaviors. For instance, a pig may simultaneously engage in feeding and standing behaviors, and in such cases, we focus on modeling the spatiotemporal relationship between the pig and the feeding trough. Finally, the specific behavior definitions and corresponding recent studies are shown in Table 1.

**Table 1**

The specific definitions of the 13 welfare behaviors and their corresponding studies; - - - behaviors annotated by domain experts in this study.

| Behavior | Behavior description | Reference | The number of annotations |
|---|---|---|---|
| drink | The pig's head is close to the nipple of the water dispenser | 25,26,27,28,31,35 | 130 |
| eat | The pig's head enters the a 3-space feeder | 30,31,32,35 | 741 |
| lying | The pig's abdomen is horizontally exposed to ground and its limbs are extended | 30,31,32,33,34,35 | 873 |
| sitting | Bending of pig forelimbs or limbs and grounding of abdomen | 33 | 3782 |
| stand | The pig's limbs extended and stood without significant displacement of its body | 30,31,33,34 | 2527 |
| move | Move within the dimension box | 31 | 1480 |
| walk | Move widely away from the dimension box | - - - | 234 |
| investigating | The pig's nose moves high-frequency towards a certain area | 34 | 2562 |
| playwithtoy | The pig's head collided violently with the concentration device | - - - | 56 |
| fight | A pig collides or moves violently with another pig | 36 | 420 |
| nose-touch-pig | The pig's nose moves high-frequency towards the other pig | - - - | 475 |
| Stand up | From lying down to sitting or standing, from sitting to standing | - - - | 106 |
| Lie down | From standing to sitting or lying, from sitting to lying | - - - | 114 |

  It is worth noting that, in addition to common behaviors such as drinking, eating, standing, sitting, moving, we have also annotated additional behaviors such as Lie down, Stand up, Nose-touch-pig, Playwithtoy, and Walk to more comprehensively assess the health and welfare of pigs. The annotation of these behaviors is of significant importance. For example, Lie down and Stand up are key indicators of pig comfort and health in the farming environment. The frequency and manner of lying down and standing up can reflect the comfort of the environment, the degree of crowding, or the pig's stress level. Specifically, the lying down and standing up behavior of sows during lactation directly impacts the safety and survival rate of piglets. Therefore, monitoring these behaviors helps optimize environmental design and reduce stress responses caused by discomfort or insufficient space.Additionally, the Nose-touch-pig behavior is an important form of social interaction among pigs, typically used for exploration, identifying companions, or establishing social connections. The frequency of this behavior can reflect the quality of the pig's social environment. A lack of nose-touch behavior may indicate isolation or stress, making the annotation of this behavior crucial for evaluating group social dynamics and individual welfare. The Playwithtoy behavior is an important way for pigs to express natural behavior and alleviate boredom, especially under confined conditions. Annotating toy-related behaviors helps measure the effectiveness of environmental enrichment and assess the pig's mental health. High-frequency play behavior is often closely related to lower stress levels and reduced behavioral abnormalities, making it an important reference for optimizing the farming environment.Finally, the Walk behavior is a fundamental part of the pig's daily activity and reflects its mobility and health status. Particularly in group activities, monitoring walking behavior can reveal the pig's level of free movement and identify potential mobility issues, such as lameness. By annotating these behaviors, we can more comprehensively assess the pig's physical activity and health status, providing data support for improving the farming environment. These additional annotated behaviors not only expand the coverage of the dataset but also provide a more comprehensive basis for multidimensional assessments of pig health and welfare, laying the foundation for optimizing scientific research and management practices.

## 2.2 Data annotation

To create a high-quality pig behavior dataset, we first preprocessed the raw video data by splitting the video into 1200-second segments, with a frame rate of 30 frames per second. We annotated the behaviors of each pig in the video segments at a rate of once per second. We used a tool called VGG Image Annotator to label the samples of pigs that needed behavior classification in the video frames, generating the corresponding label files. The data that needed manual annotation included the coordinates of the four corners of the rectangular area around the pig, as well as the pig's ID and its behavior within that area. Figure 2 shows the interface of the annotation software and the process of recording annotations. "Category" represents the ID of each pig, "scores" refers to the prediction score of the pig detector, and "behavior" indicates the behavior of the pig.

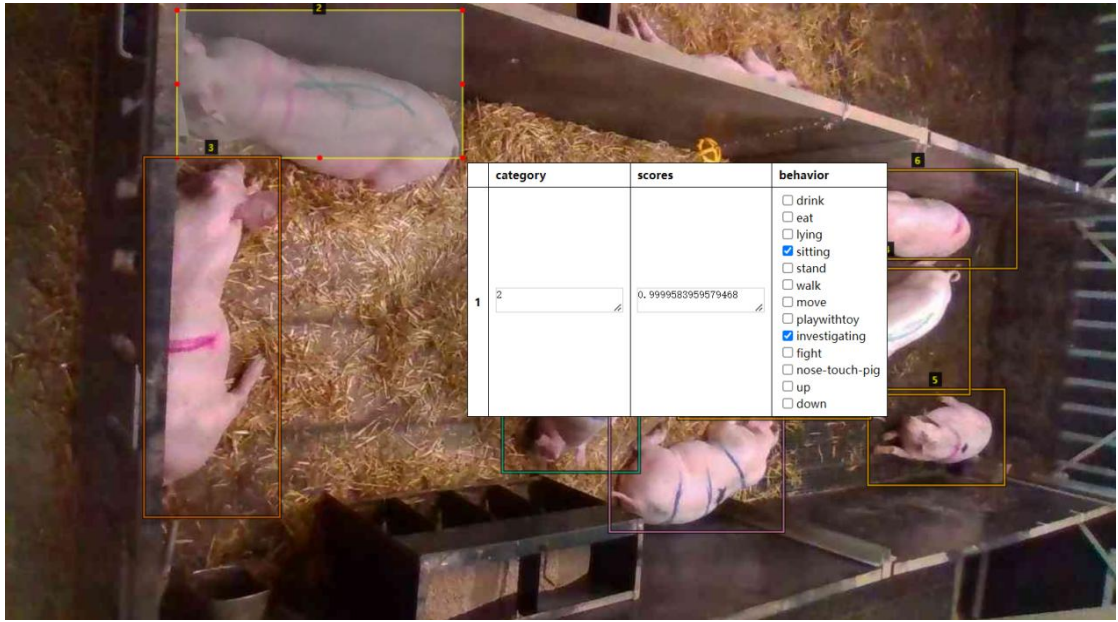

**Figure 2.** Using the VGG Image Annotator annotation process

Each video frame contains eight pigs with special markings on their backs. In actual farming situations, the activities of pigs are very complex. In the captured video frames, pigs may be obscured by other pigs. This situation is detrimental to behavior recognition(zhang et al.2019[37]). Therefore, to avoid incomplete spatiotemporal feature extraction caused by occlusion, we define a pig as "Hidden" if it is obscured by more than two-thirds in either the spatial or temporal dimension. An example is shown in Figure 3. In the given video frame sequence, although the pig labeled in the last frame is not heavily occluded in the spatial dimension, it is almost completely covered in the temporal dimension. In this case, the model is unable to extract the features required for detecting the pig's sitting behavior.

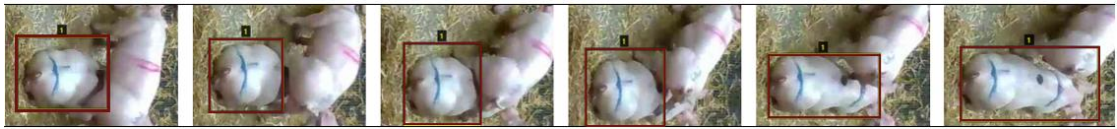

**Figure 3**.Example of pigs obscured by 2/3 in temporal and spatial dimensions.

In this study, the preprocessed dataset was divided into training, validation, and test sets according to a 7:2:1 ratio. Finally, based on the organization structure of the AVA dataset, we annotated nearly 19,200 bounding boxes for the 8 pigs appearing in the 1200-second video segments.

## 2.3 Research methodology

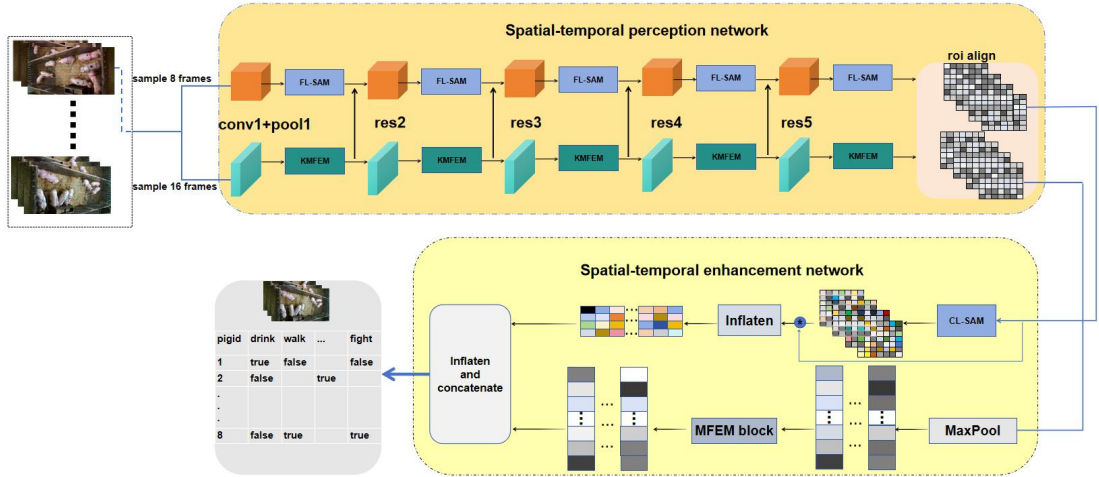

**Figure 4.**The network structure diagram, where FL-SAM and KMFEM represent the Frame-level Spatial Attention Module and Key Motion Feature Extraction Module, respectively. CL-SAM and MFEM represent the Frame-level Spatial Attention Module and Motion Feature Enhancement Module, respectively. The orange and cyan blocks are the residual blocks that form the network.

The overall framework of the network is shown in Figure 4. The input to the model is two sequences of pig behavior video frames derived from the same video using different sampling rates. The model processes these sequences through two network branches.The models are divided into the spatiotemporal perception network and the spatiotemporal feature enhancement network. The spatiotemporal perception network is responsible, using its two branches, for extracting the key spatial features related to the target pig and their variations in these features. Then, using the pre-annotated and normalized bounding boxes, network crop the feature maps from both branches to obtain the key feature maps related to each pig and its behavior. Finally, in the Spatiotemporal Feature Enhancement Network, we enhance the spatial features of individual pig behavior sequences using the channel-level spatial attention module and model the spatiotemporal features of pig behaviors using the motion feature enhancement module. The results from these two branches are then concatenated and fused to obtain the feature vectors for each pig in the video data, which are then fed into the classification head to achieve behavior classification.

## 2.3.1 Spatiotemporal perception network

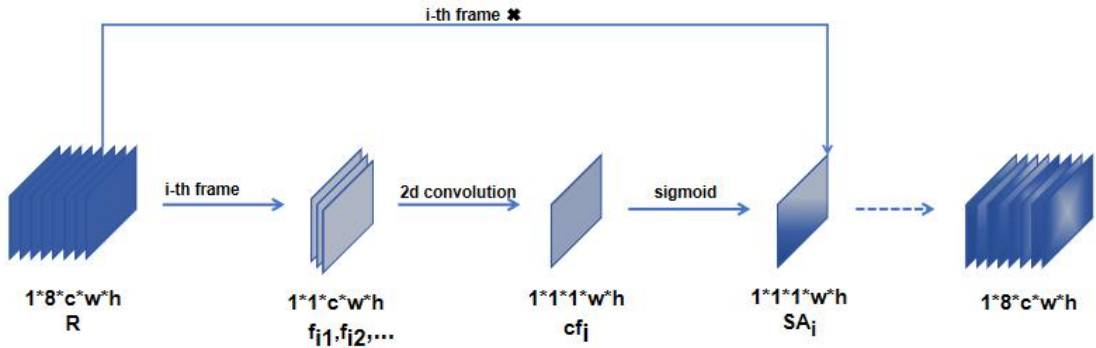

**Figure 5.**Internal structure diagram of FL-SAM

The spatiotemporal perception network consists of the spatial attention residual network and the interframe difference residue network. The spatial attention residual network includes five residual blocks, with a total of four Frame-Level Spatial Attention Modules (FL-SAM) inserted between adjacent residual blocks. The branch is responsible for extracting the key spatial features of individual pig behavior in the low-frame-rate branch. In the inter-frame difference residual network, the five residue blocks and the Key motor feature extraction module(KMFEM) inserted between adjacent residual blocks is responsible for extracting the changes in key spatial features in high frame rate branches. after extracting the key spatial features of individual pig behavior and the changes in key spatial features, the two branch will fuse at the low frame rate branch by adding. Finally, ROI is used to crop the feature maps from both branches to obtain the spatial features of relevant positions and their motion variations for each pig during movement.

The Frame-Level Spatial Attention Module (FL-SAM) assigns different computational weights to the extracted spatial features of pig behaviors from different frames during the model's feature extraction process. This approach enhances the model's ability to represent the key space of the specific action of the pig behavior. For instance, as shown in Figure 5, the shape of the frame sequence feature map is batch, time, channel, width, height, with each video frame initially having c channels. Assuming we have a sequence sample V={f1,f2,...,f8} obtained from the sequence sampling, the feature maps for video frame sequence after passing through the residual blocks are denoted as R={{$f_{11}$,$f_{12}$,...},{$f_{21}$,$f_{22}$,...},...,{$f_{81}$,$f_{82}$,...}}, where $f_{ij}$ is the feature of the j-th channel in frame i. Since 2D convolution operates through a local receptive field, it can capture the correlations between different channels or local spatial patterns. Additionally, by compressing the information from C channels into a single channel through convolution, it can generate a global summary of all spatiotemporal features. Subsequently, within the FL-SAM block, the model performs further operations on each frame's corresponding feature maps in R using 2D convolution, yielding CR={{$cf_1$},{$cf_2$},...,{$cf_8$}}, where the channel count is reduced to 1. Finally, the weight feature map SA is computed using Equation (1).

$$SA = \{sigmoid(cf_1), sigmoid(cf_2),...,sigmoid(cf_8)\} \quad (1)$$

Finally, by Formula (2), the feature maps CR and SA are multiplied element-wise to select the spatial features L that have a significant impact on the classification of pig welfare behaviors.

$$L = \{\{f_{11}, f_{12},...,\},\{f_{21}, f_{22},...\},\{f_{i1}, f_{i2},...,\}\} \times sigmoid(cf_i), i \in [1,2,...,8] \quad (2)$$

In the inter-frame difference residual network, the spatial involved in detecting the target's motion typically occupies only a small portion of the labeled bounding box. However, traditional methods that extract motion features through inter-frame differencing not only capture spatial information related to the target but often also include other unrelated targets, which adversely affects the extraction of motion features.To address this issue, we propose a new Key motor feature extraction module(KMFEM) that focuses the model on the spatial changes relevant to the detected target. This module first employs a spatial attention mechanism to concentrate the model's focus on spatial areas that significantly impact pig welfare behaviors, following the same approach as FL-SAM. Subsequently, it computes the inter-frame differences in the relevant spatial areas to obtain the motion feature flow. The pseudocode for this process is as follows:

Algorithm  Pseudocode for key motion feature extraction

```
# Input:
# frames - List of high-frame-rate video frame features [f1, f2, ..., f16]
# weights - Attention weights for the frames
# Output:
# feature - Key spatial feature map after applying attention
# motion - Motion-sensitive feature map based on frame differences
# convolved_motion - Further refined motion features using convolution

# Step 1: Compute attention weights for the frames
weights = compute_attention(frames)    # Compute attention weights for each frame

# Step 2: Apply attention weights to the frame features
feature = frames * weights    # Element-wise multiplication of frames and weights

# Step 3: Split sequence into front and back sequences for motion calculation
front15 = frames[:-1]    # All frames except the lase one
back15 = frames[1:]      # All frames except the first one

# Step 4: Compute motion-sensitive feature map
motion = back15 - front15 # Element-wise difference between consecutive frames

Step 5: Refine motion-sensitive features using convolution
convolved_motion = convolve(motion)    # Apply convolution to enhance motion features

# Return results
return feature, motion, convolved_motion
```

### 2.3.2  Spatiotemporal feature enhancement network

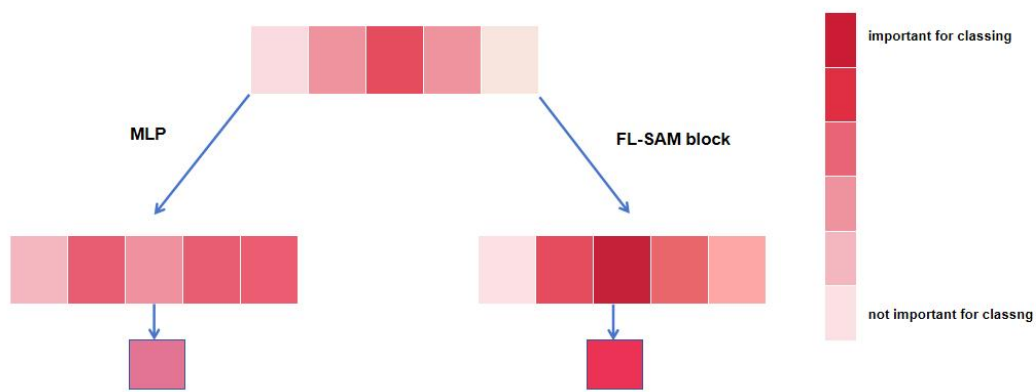

**Figure 6.** Schematic representation of the feature calculation performed by MLP and CL-SAM

In the spatiotemporal feature enhancement network shown in Figure 4, based on the key spatial features and subtle motion features of individual pigs extracted on the ROI Align module, first in low frame rate branch using the channel level attention mechanism module (CL-SAM) for individual pig behavior sequence spatial feature enhancement, in addition for the change of pig movement features,this paper use the motion feature enhancement module (MFEM) to remodel the temporal features of pig behavior, and then the results of the two branches leveling fusion, the final classification of pig behavior.

The frame-level spatial attention mechanism effectively extracts spatial features from the key parts of pigs' movement processes. Traditional network models then directly pool, compute with MLP, and classify these features. However, features extracted by deep networks vary in their representational power for pig welfare behaviors, meaning not all feature maps contribute equally to behavior classification. As shown in Figure 6, while conventional MLP algorithms can significantly extract key features for behavior classification, other feature maps with weaker representational power still have some impact on classification. To minimize similar negative impacts, this paper uses a channel-level spatial attention mechanism (CL-SAM) in the low-frame-rate branch to re-weight each pig's feature map sequence obtained via ROI Align, enhancing features that play a crucial role in behavior classification. This approach allows the network to better focus on identifying and understanding behavior-related details, thus improving the model's comprehension of video content. This is achieved by applying a sigmoid function to the post-ROI feature maps to restrict values between 0 and 1, followed by element-wise multiplication with the original feature maps.

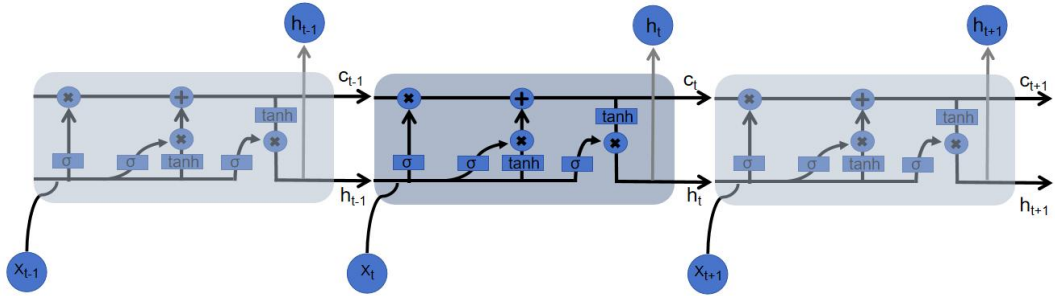

**Figure 7**. Structural diagram of the LSTM

The motion patterns in pig behavior have a significant impact on classifying pig welfare behaviors. For instance, standing up and sitting down are opposite sequences in the temporal dimension; however, conventional models often classify such behaviors with MLP, disregarding this temporal information, which is detrimental to accurate pig behavior classification. Therefore, we propose a Motion Feature Enhancement Module (MFEM) to strengthen the temporal information in pig behaviors. In this study, we employ LSTM to model the changes in pig motion, and the LSTM structure is shown in Figure 7.In the 16-frame branch of the network, after processing through the inter-frame difference residual network, the network uses ROI Align to obtain a feature sequence for each pig in the video. Then, by applying max-pooling, it selects the most impactful features from each feature map concerning pig movement behavior, forming a time sequence that is fed into the LSTM for temporal modeling. LSTM utilizes its gating mechanism to retain information relevant to motion behavior while discarding unnecessary details, helping the model to understand pig behavior along the temporal dimension.

## 3. Experiment and Analysis

### 3.1 Experimental environment and hyperparameters

The experiments were conducted on a computer configured with 64GB RAM, an NVIDIA GeForce RTX 4090 GPU with 24GB of memory, and a 64-bit Ubuntu system. Anaconda3 was used as the support platform, Jupyter as the integrated development environment, and PyTorch as the inference framework. We selected TSN, TSM, I3D, SlowFast and ACRN as comparison models in this study.The hyperparameters for all experiments were uniformly set as follows: the batch size was 32; the input image sequence size for the low-frame-rate branch was 8×224×224, and for the high-frame-rate branch, it was 16×224×224. The optimizer used was SGD, and the learning rate was adjusted automatically during training. TensorBoard was utilized to collect various metrics during the network training process. The initial learning rate was set to 0.001, with a weight decay of 0.02. A cosine annealing learning rate decay strategy was adopted, where the learning rate was annealed from the initial value to the minimum value based on a cosine function within each stage from 0 to 60, and then restarted in the subsequent stage.

### 3.2 Evaluation metric

In this study, we use Average Precision (AP) as the fundamental evaluation metric for pig behavior recognition. Pig behavior recognition is a multi-label classification task. By inferring scores for each behavior of each pig through the proposed network model, we can understand the model's performance across different action categories. The AP metric for each pig behavior provides insights into the model's classification performance across various categories, particularly for imbalanced behaviors in the dataset.AP is derived from commonly used evaluation metrics in multi-label classification tasks: recall and precision. Precision measures the proportion of true positives among the predicted positives, reflecting the accuracy of the model's predictions. Recall measures the proportion of correctly identified positives among the actual positive samples, reflecting the model's sensitivity to positive samples. Finally, we use the mAP metric to evaluate the overall performance of the model. The formulas for Recall, Precision, and mAP are as follows:

$$P = \frac{TP}{TP+FP} \tag{3}$$

$$R = \frac{TP}{TP+FN} \tag{4}$$

$$AP_i = \sum_{j=2}^{n}(R(j)-R(j-1))\times P(j) \tag{5}$$

$$MAP = \frac{\sum_{i=1}^{k}AP_i}{k} \tag{6}$$

In formulas (3) and (4), TP, FP, and FN are calculated based on behavior prediction scores. In formula (5), R represents Recall, P represents Precision, and R(n),R(n−1),⋯,R(2),R(1) are arranged in descending order.

## 3.3 Model Comparative experiment

This study conducted comparative experiments between the proposed model and traditional spatiotemporal feature extraction models, including TSN, TSM, I3D, and SlowFast. Additionally, to compare the spatiotemporal modeling capability of the proposed model, ACRN was also selected as a benchmark model. All specific behaviors were evaluated using Average Precision (AP) as the performance metric.

As shown in Table 2, the proposed model achieved a Mean Average Precision (mAP) on the 13-behavior dataset that outperformed the TSN, TSM, I3D, and SlowFast models by 14.22 percentage points, 14.94 percentage points, 14.68 percentage points, and 9.41 percentage points, respectively.

Table 2.
The behavior classification algorithm proposed in this paper and the six comparative models were evaluated on the dataset established in this study. The average precision for each behavior and the Mean Average Precision (mAP) of each model are presented.

|  | TSN | TSM | I3D | SlowFast | ACRN | Ours |
|---|---|---|---|---|---|---|
| drink | 70.9% | 96.7% | 70.4% | 81.28% | 83.5% | 81.63% |
| eat | 98.24% | 59.52% | 96.57% | 95.06% | 96% | 95.75% |
| lying | 76.27% | 77.15% | 91.71% | 74.37% | 85.22% | 96.67% |
| sitting | 80.58% | 81.71% | 89.03% | 82.45% | 80.75% | 94.23% |
| stand | 93.2% | 93.06% | 92.63% | 92.24% | 91.9% | 92.29% |
| move | 72.64% | 80.41% | 80.42% | 81.75% | 81.51% | 82.77% |
| walk | 78.32% | 78.76% | 81.26% | 82.28% | 79.28% | 90.13% |
| investigating | 61.93% | 64.51% | 61.29% | 66.28% | 66.28% | 72.48% |
| playwithtoy | 85.47% | 78.93% | 19.43% | 77.07% | 80.73% | 85.33% |
| fight | 29.01% | 30.22% | 37.33% | 36.92% | 40.92% | 60.75% |
| nose-touch-pig | 15.1% | 11.92% | 22.08% | 36.53% | 39.21% | 40% |
| Stand up | 24.23% | 25.18% | 33.52% | 38.36% | 30.56% | 52.37% |
| Lie down | 16.2% | 14.72% | 20.62% | 19.54% | 23.54% | 41.98% |
| MAP | 61.7% | 60.98% | 61.24% | 66.51% | 67.75% | 75.92% |

This paper adopts a spatiotemporal perception and enhancement network based on the attention mechanism to eliminate redundant spatiotemporal information in video data for individual pig behavior recognition. By enhancing and modeling the key spatiotemporal features of individual pigs, it efficiently captures the sensitive features related to pig behavior recognition in the video. Compared with other traditional models for video data behavior recognition, this model demonstrates higher recognition accuracy. Notably, the improvement in behavior classification, especially for time-dependent behaviors such as Stand Up and Lie Down, is particularly significant.

Taking the ACRN model as an example for comparative analysis, the ACRN model consists of two stages: object detection and action classification. Since the dataset proposed in this paper has already completed the labeling of pig bounding boxes, there is no need for object detection, so we focus more on its action classification principle. ACRN obtains the subject's feature map through ROI and performs a spatial correlation operation with the global feature map to model the spatiotemporal relationship between the pig and other regions within its activity range. The result is then concatenated with the subject's feature map and classified. Experimental results show that, except for the model proposed in this paper, ACRN demonstrates a significant improvement in MAP compared to other models, proving that modeling the relationship between the behavior subject (the pig) and the environment can significantly enhance model performance.The pig behavior recognition algorithm proposed in this paper approaches the problem from a different angle by modeling the pig (as the subject) and its related behavioral activity regions. First, the spatiotemporal perception network uses FL-SAM in the high and low frame rate branches, allowing the network to focus more closely on the important regions related to pig behavior in the video. Additionally, the KMFEM in the high frame rate branch calculates the changes in key regions. As the depth of the model increases, the receptive field also expands. Ultimately, by using ROI operations, we obtain not only the subject feature map of the pig but also the key region features and their changes, significantly improving the accuracy of space-sensitive behaviors such as nose-touch-pig, lying, and sitting.Furthermore, the CL-SAM in the spatiotemporal enhancement network further evaluates the spatial features of the moving subject, while the MFEM remaps the pig's motion changes, greatly enhancing the model's recognition precision for time-sequence-related behaviors such as fight, investigating, stand up, lie down, and walk.

## 3.4 Ablation experiment

This experiment conducts an ablation study on the FL-SAM and KMFEM in the low and high frame rate branches of the spatiotemporal perception network, as well as the CL-SAM and MFEM in the spatiotemporal feature enhancement network. The results are compared with the classification results of the entire model proposed in this paper across various metrics. The controlled variable method is used, where all hyperparameters remain consistent except for the two different input schemes. The metric results are shown in Table 3.

Table 3.
The ablation study of the four modules in the model proposed in this paper is conducted to evaluate their individual contributions to the overall performance.

|  | Whole | No FL-SAM | No KMFEM | No CL-SAM | No MFEM |
|---|---|---|---|---|---|
| drink | 81.63% | 77.42% | 77.55% | 76.37% | 84.45% |
| eat | 95.75% | 96.31% | 97.15% | 96.76% | 95.97% |
| lying | 96.67% | 94.49% | 94.85% | 93.83% | 95.41% |
| sitting | 94.23% | 92.15% | 92.59% | 90.82% | 90.61% |
| stand | 92.29% | 94.93% | 94.69% | 94.02% | 94.79% |
| move | 82.77% | 80.14% | 81.31% | 80.18% | 80.94% |
| walk | 90.13% | 86.37% | 81.44% | 83.39% | 84.02% |
| investigating | 72.48% | 65.34% | 75.54% | 69.47% | 71.83% |
| playwithtoy | 85.33% | 83.54% | 82.55% | 86.20% | 96.67% |
| fight | 60.75% | 64.53% | 56.79% | 44.84% | 50.25% |
| nose-touch-pig | 40% | 33.11% | 36.45% | 33.43% | 28.63% |
| Stand up | 52.37% | 44.99% | 49.67% | 55.20% | 39.37% |
| Lie down | 41.98% | 40.30% | 25.92% | 36.42% | 23.29% |
| MAP | 75.92% | 73.36% | 72.81% | 72.38% | 72.02% |

The ablation experiment results indicate that the FL-SAM, KMFEM, CL-SAM, and MFEM modules each play an indispensable role in improving the model's performance. The introduction of FL-SAM significantly enhanced the model's ability to classify behaviors dependent on spatial features. Without FL-SAM, the classification accuracy for behaviors such as "drink," "lying," and "sitting" dropped to 77.42%, 94.49%, and 92.15%, respectively, compared to the complete model's 81.63%, 96.67%, and 94.23%. This demonstrates that FL-SAM effectively extracts the key spatiotemporal features of pig movement, enhancing the model's ability to classify behaviors sensitive to spatial features, while also improving the overall model's generalization and accuracy. KMFEM plays a key role in extracting temporal features in the spatiotemporal perception network. Without KMFEM, the classification accuracy for behaviors dependent on temporal features, such as "move" and "walk," dropped to 81.31% and 81.44%, while the accuracy for the complete model was 82.77% and 90.13%. This indicates that KMFEM, through the frame-differencing method, more comprehensively extracts motion change features in key areas, significantly enhancing the representation of temporal features in behavior changes.The CL-SAM module further improves the model's classification accuracy by focusing on key parts of the pig's spatial feature map. The experimental results show that after adding the CL-SAM module, the model's MAP increased from 72.38% to 75.92%, an improvement of 3.54%. Specifically, the accuracy for the "lying" behavior increased from 93.83% to 96.67%, a 2.84% improvement, while the accuracy for the "investigating" behavior rose from 69.47% to 72.48%, a 3.01% improvement. This indicates that the CL-SAM module effectively highlights key spatial features, enhancing the model's performance in spatial feature extraction.The MFEM module, by introducing an LSTM model for temporal feature modeling, significantly improves the model's sensitivity to changes in the temporal dimension. The experimental results show that without the MFEM module, the accuracy for the "Stand up" behavior dropped from 52.37% to 39.37%, and the accuracy for the "Lie down" behavior decreased from 41.98% to 23.29%. Other behaviors, except for "drink," "eat," and "stand," also experienced slight accuracy drops. These results indicate that the MFEM module enhances the model's understanding of temporal changes by modeling key features over time, showing clear advantages, especially when handling behaviors that rely on temporal features.

Table 4.

Combined experiments of the four modules.

| | Fl-SAM+KMFEM | FL-SAM+CL-SAM | FL-SAM+MFEM | KMFEM+CL-SAM | KMFEM+MFEM | CL-SAM+MFEM |
|---|---|---|---|---|---|---|
| MAP | 70.11% | 69.54% | 72.19% | 69.7% | 71.87% | 70.6% |

Finally, this study examines the combined performance of the four modules (FL-SAM, KMFEM, CL-SAM, and MFEM) by comparing their mAP metrics to analyze the synergistic effects between different modules and their impact on behavior classification tasks. The experimental results are shown in Table 4. From the table, it can be observed that the model's performance is relatively higher when using MFEM. However, in the combination experiment of KMFEM and MFEM, even without CL-SAM, the accuracy is only 0.98% lower than when CL-SAM is used. This suggests that in the spatiotemporal perception network, when performing ROI operations on the feature maps, the receptive field of the cropped subject feature map increases with the depth of the model, and implicitly incorporates the surrounding environment of the pig. This implicitly establishes the relationship between the behavior classification subject and the surrounding environment.Additionally, in the experiments combining FL-SAM with CL-SAM and KMFEM with CL-SAM, the results without FL-SAM also support this hypothesis. However, the mAP in these two experiments is still lower than the combination results that include FL-SAM, further proving that FL-SAM plays an irreplaceable role in the complete model.

### 3.5 Discussion

In traditional farm animal behavior recognition, particularly in large-scale pig farming, the behavior recognition typically involves identifying multiple targets. Traditional spatiotemporal action classification models extract features for classification that include all spatial data across dimensions. However, in large-scale farm settings, the features relevant to accurate behavior prediction are often only related to a small portion of the space surrounding the target. As a result, traditional action detection models tend to focus on too many irrelevant features, which reduces the model's performance.

In this study, we employ a spatiotemporal perception network to establish a connection between the behavior classification subject (the pig) and key areas of the surrounding environment. We also use a spatiotemporal enhancement network to re-model the spatial and temporal features of each pig, ultimately achieving the highest recognition accuracy. However, in the experiments shown in Table 3, we observed that when MFEM was not used, behaviors closely related to spatial features, such as "drink," "eat," and "play with toy," reached the highest accuracy in the ablation experiments. We hypothesize that as the model's receptive field increases with its depth, in less complex behavior classification scenarios, the model may also capture the relationship between the behavior subject and the surrounding environment.However, without using FL-SAM, the model performance noticeably decreased, further proving the importance of regions related to the behavior subject. The effectiveness of FL-SAM is also dependent on the convolutional receptive field being able to capture these regions. Therefore, in pig behavior classification tasks, it is crucial to find an algorithm that explicitly models the relationship between the action executor and the key regions of the action.These questions will be further explored in our future work.Figure 8 presents attention heatmaps for the "eat" and "lie down" behaviors, analyzing the model's effectiveness. When the pig is performing the "eat" behavior, the model focuses its attention on the pig's head and the area where the head contacts the feeding trough. For the "lie down" behavior, the model focuses on the pig's abdomen and hindquarters, where noticeable displacement occurs.

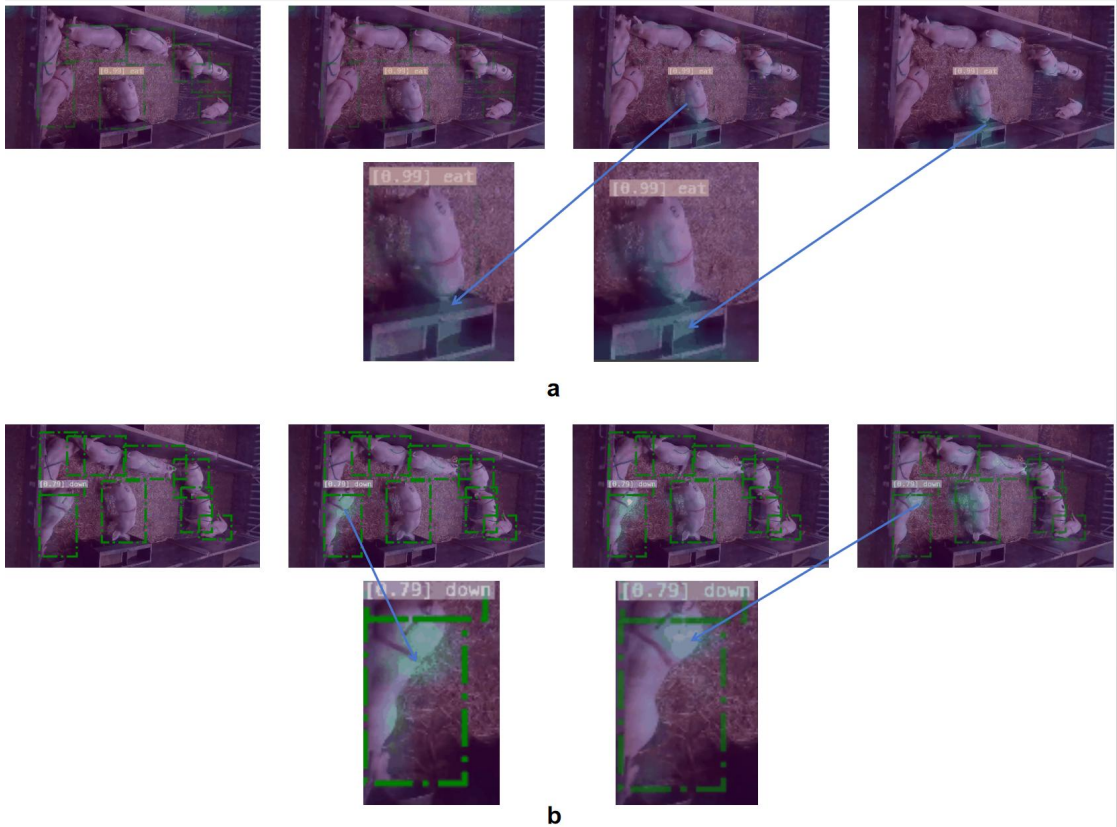

Figure 8. A and b are the visual samples of ease and lie down of the proposed network.

The dataset established in this study not only covers pig behaviors that have been widely researched in recent years but also includes additional behaviors such as "Lie down," "Stand up," "Nose-touch-pig," "Play with toy," and "Walk," which have significant impacts on pig welfare. Compared to traditional models, the proposed model not only achieves excellent performance but also successfully models the key spatiotemporal features of individual pigs in complex video data. This approach improves the recognition rate of pig behavior classification and enhances the model's generalization ability and accuracy.

## 4. Conclusion

This study proposes and publicly releases a dataset containing 13 pig behaviors that have significant impacts on pig welfare. Based on this dataset, we introduce a spatiotemporal perception and enhancement network based on attention mechanisms, which models and recognizes the key spatiotemporal features related to individual pig behaviors in pig video data. The spatiotemporal perception network successfully establishes connections between the pigs and the key regions of their behaviors in the video data, while the spatiotemporal enhancement network further perceives and enhances the spatial and temporal features of pig behaviors, re-modeling the spatiotemporal features of the pig as the subject. Finally, through comparative and ablation experiments, the results demonstrate that the proposed model achieves an mAP score of 75.92% on the dataset we established, improving by 8.17% over the best-performing traditional models.

**Declaration of Competing Interest**

The authors declare that they have no known competing financial interests or personal relationships that could have appeared to influence the work reported in this paper.

## Data availability

Data will be made available on request.

## Acknowledgement

Thanks for the support of the College of Computer and Artificial Intelligence, Changzhou University.This study did not receive any specific funding from funding organisations in the public, commercial or not-for-profit sectors。